\documentclass[twocolumn]{article}

\usepackage{arxiv}
\usepackage{multirow}
\usepackage[utf8]{inputenc} % allow utf-8 input
\usepackage[T1]{fontenc}    % use 8-bit T1 fonts
\usepackage{hyperref}       % hyperlinks
\usepackage{url}            % simple URL typesetting
\usepackage{booktabs}       % professional-quality tables
\usepackage{amsfonts}       % blackboard math symbols
\usepackage{nicefrac}       % compact symbols for 1/2, etc.
\usepackage{microtype}      % microtypography
\usepackage{lipsum}
\usepackage{graphicx}
\usepackage{algorithm}
\usepackage{algpseudocode}
\usepackage{times}
\usepackage{epsfig}
\usepackage{amsmath}
\usepackage{amssymb}
\usepackage[compact]{titlesec}         % you need this package
\graphicspath{ {./images/} }

\titlespacing{\section}{0pt}{0pt}{0pt} % this reduces space between (sub)sections to 0pt, for example
\AtBeginDocument{%                     % this will reduce spaces between parts (above and below) of texts within a (sub)section to 0pt, for example - like between an 'eqnarray' and text
  \setlength\abovedisplayskip{0pt}
  \setlength\belowdisplayskip{0pt}}

\begin{document}

%%%%%%%%% TITLE
\title{Pose2RGBD. Generating Depth and RGB images from absolute positions.}

\author{Pîrvu Mihai Cristian\\
University "Politehnica" of Bucharest \& MorphL\\
{\tt\small mihaicristianpirvu@gmail.com}
}

\twocolumn[
\begin{@twocolumnfalse}
\maketitle
%%%%%%%%% ABSTRACT
\begin{abstract}
	We propose a method at the intersection of Computer Vision and Computer Graphics fields, which automatically generates RGBD images using neural networks, based on previously seen and synchronized video, depth and pose signals. Since the models must be able to reconstruct both texture (RGB) and structure (Depth), it creates an implicit representation of the scene, as opposed to explicit ones, such as meshes or point clouds. The process can be thought of as neural rendering, where we obtain a function $f : Pose \rightarrow RGBD$, which we can use to navigate through the generated scene, similarly to graphics simulations. We introduce two new datasets, one based on synthetic data with full ground truth information, while the other one being recorded from a drone flight in an university campus, using only video and GPS signals. Finally, we propose a fully unsupervised method of generating datasets from videos alone, in order to train the Pose2RGBD networks. Code and datasets are available at: \url{https://gitlab.com/mihaicristianpirvu/pose2rgbd}.
\end{abstract}
\end{@twocolumnfalse}
]

%%%%%%%%% BODY TEXT
\section{Introduction}

The field of Computer Vision has been mostly concerned with inverse graphics problems, which takes one or more pictures and tries to explain them using different levels of representations, such as classification \cite{deng2009imagenet}, depth estimation \cite{eigen2014depth, marcu2018safeuav}, object detection \cite{redmon2018yolov3} or semantic segmentation \cite{Cordts2016Cityscapes}. On the other hand, the field of Computer Graphics has mostly focused on generating fast and high fidelity sceneries using handcrafted models. Their main focus is either realism, with advances both in level of detail for the used models and algorithmic, such as ray-tracing \cite{suffern2016ray} or speed, using various engineering and hardware solutions that enables the solutions for consumer level PCs.

Classical rendering pipelines involves processing hard-coded meshes, texture mapping, applying lighting, shadows and other effects, while simultaneously being conscious about clipping and level of detail of far or occluded views in order to minimize the amount of work that needs to be done on GPU. Then, the system must also map 3D coordinates to a 2D viewport, and this process is repeated at every frame, with very little room for parallelism ahead of time.

% Differentiable renderers and pixel2mesh.
Recently, there has been some traction in the domain of differentiable rendering \cite{li2018differentiable, palazzi2018end, kaolin2019arxiv}, which proposes end-to-end differentiable methods of simulating the traditional rendering pipelines, processing explicit representations, such as meshes, to produce 2D projections and texture mappings. Inverse problems have also been tackled, like generating meshes from RGB images \cite{wang2018pixel2mesh}. We come to complete the picture by generating novel views from absolute positions directly, bypassing any sort of dense and redundant representations of a view.

Classical Computer Vision pipelines, based on the idea of Structure from Motion \cite{klein2007parallel} and Visual Odoemtry \cite{nister2004visual} try to solve the same problem, which is representing a 3D model acquired using 2D imagery that are tied together using various algorithms. A typical pipeline involves gathering some images of a scene, using various algorithms to extract keypoints in each image \cite{lowe2004distinctive}, merging them together, by predicting a relative pose between the images, and then completing the 3D model image by image. Usually the model is stored explicitly as point clouds, which may also impose memory restrictions. This process is very error prone, but has been a main research topic and has led to very impressive results. Having an explicit 3D model is not always practical, and thus, we try to solve this issue by using a neural network to implicitly memorize the 3D structure and directly infer the 2D projections when required at any given pose inside some predefined boundaries.

Other closely tied problems are the tasks of inferring the absolute pose from imagery \cite{kendall2015posenet}, navigating through novel scenes \cite{henriques2018mapnet}, predicting dense maps depth maps from RGB inputs in supervised \cite{eigen2014depth} or unsupervised fashion \cite{zhou2017unsupervised}, as well as generating novel RGB images from noise using generative models \cite{kingma2013auto, goodfellow2014generative}.

Finally, the closest article related to our work is \cite{eslami2018neural}, which uses RGB images and absolute poses in order to encode information about a specified scene. Their Generative Query Network is then able to implicitly render novel views in their simulated environment. Unlike their work, we show that the model can be trained using only absolute pose and can infer both RGB and Depth images simultaneously. We also use much more complicated datasets, that simulates a flying UAV in an outdoor scene.

While this work is not directly aimed at completely replacing the traditional rendering mechanism, it aims to offer new way of producing 2D projections from an implicit 3D model, using a neural network that regresses both textures and structure in the forms of RGB and Depth images, from an absolute position. The stored 3D scene is just an approximation of the original scene, so the level of detail will most likely be reduced, however, in most cases, not every intricate detail is needed to make high level inferences, like avoiding obstacles or detecting objects. The main difference between the classical rendering pipeline and this work is that the details of the world are compressed in a high level representation directly from observed data. The network incorporates in its weights a compressed implicit 3D model, which can then be referenced and used to produce novel viewpoints.

In the following section, we'll discuss about our proposed method and network architecture. Then, we'll introduce the synthetic and real world datasets that were used followed by a general technique to create a new dataset from scratch for this problem using only a video footage. Finally, we'll provide a series of results and studies we ran, followed by discussing possible future directions and improvements.

\section{Proposed method}
\label{sec:proposed-method}
%- talk about the general method

The proposed model, succinctly called Pose2RGBD, aims to compute a function $f : (P, w) \rightarrow RGBD $, represented as a neural network, that takes as input an absolute pose (P) and makes a dense estimation of a 4 channel map (RGBD), by adjusting the trainable weights (w) such that a pixel-wise cost function is minimized between the predictions and the ground truth values of the training set. The absolute pose is represented as 6DoF (translation + rotation) restricted to a space of $[-1 : 1]$, by applying min-max normalization over its extremes. The translation is represented simply as the offset from the center of the surveilled space. The rotation can represented both as Euler angles ($\phi, \theta, \psi$) or quaternions ($a, b, c, d$), however the later case gave us better results, and thus was prefered.

\begin{figure*}
\begin{center}
\fbox{
\includegraphics[width=0.8\linewidth]{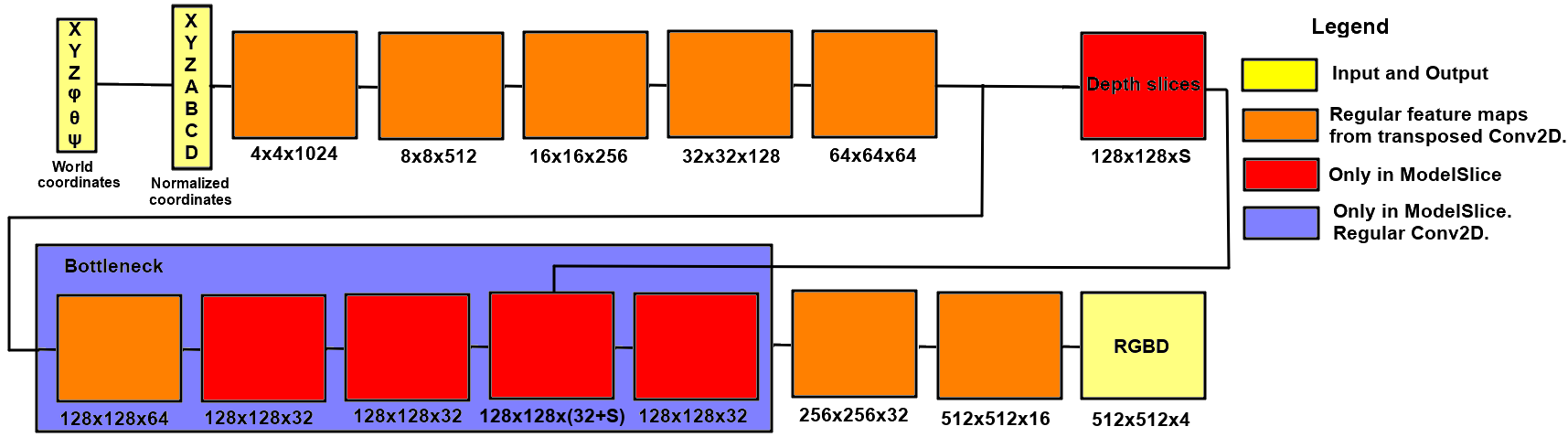}
}

\end{center}
   \caption{Pose2RGBD Model Architecture. Inputs are represented as 6DoF poses, while outputs are dense 4-channeled RGBD maps.}
\label{fig:model-architecture}
\end{figure*}

The model must then project this low dimensional and non-redundant information into a high dimensional dense representation. The output RGBD map is also normalized to $[-1 : 1]$ and then reprojected into the original space for computing metrics, such as Depth error in meters (for the synthetic dataset) or Absolute Pixel error (for both datasets). The proposed architecture is inspired by the generator of DCGAN \cite{radford2015unsupervised} and is described in Figure \ref{fig:model-architecture}.

There are two versions of models here. The first one, which contains only orange boxes, is simply the generator of the DCGAN, starting from a 6DoF representation, using a dense layer to project it to a higher dimensional space and then reshaping to 4x4x1024, in order to be processed by 2D transposed Convolution operations, until reaching a 512x512x4 output. The loss function is computed using standard pixel-wise MSE. The second version, titled Model Slice, has a few additional quirks. First, the introduction of a Bottleneck layer, which aims process the high level feature maps. All the operations done in this layer are done using traditional 2D convolutional operations, with a kernel size of 3. Then, we are also using a secondary output map, where we slice the depth map in multiple progressive layers. Basically, if our depth map has an output range of $[-1 : 1]$, then, for $S=10$ slices, we'll get 10 binary channels with positive values at the locations where the depth map is inside each slice's interval. The slices are divided uniformly, so the first one captures the depth in range $[-1, -0.8)$, the second one has the range $[-0.8, 1)$ and so on until the last one that includes the range $[0.8 : 1]$. The loss function on this layer is computed using standard binary cross entropy for each slice independently. Also, the slice are concatenated to the feature maps of the penultimate layer of the bottleneck. The reasoning for this is simply to combine the high level feature maps with the hand-crafted features caused by the layering. We've varied the number of slices, which offers a trade-off between speed and performance, as we'll present in the experiments section. Finally, these slices can be interpreted as confidence maps by summing them. Batch Normalization is used between all layers, except the output ones, where we use hyperbolic tangent for the predicted RGBD map and sigmoid for the depth slices.

\subsection{Creating a dataset from scratch}

In order to train the Pose2RGBD network, one needs to have all 3 sources of information synchronized: RGB, Absolute Pose and Depth for each frame of a video. The network can be trained on multiple videos as well, however they should all be in the same spacial environment. Having two identical poses pointing to two different images will simply make no sense.

The first case will assume that we have collected a set of videos and localization data, however they are unsynchronized and recorded at different frequencies.

\begin{algorithm}
\caption{Synchronizing RGB and Absolute Pose}
\label{alg:sync-rgb-abspose}
\begin{algorithmic}[1]
\State $ GPS' \gets \Call{interpolate}{GPS, RGB-Freq} $
\State $ Flow \gets \Call{Optical Flow}{RGB} $
\State $ Flow-Mag \gets \sqrt{Flow(u)^{2} + Flow(v)^{2}}$
\State $ GPS-Sync \gets \Call{Match}{Flow-Mag, GPS'} $
\State $ \Return RGB, GPS-Sync $
\end{algorithmic}
\end{algorithm}

\begin{figure}[h]
\begin{center}
   \includegraphics[width=1\linewidth]{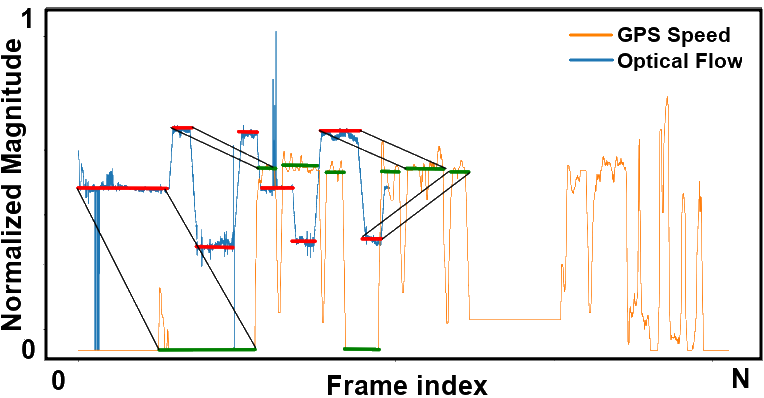}
\end{center}
   \caption{GPS Speed vs Optical Flow magnitude in a video. Manual synchronization was done in critical points.}
\label{fig:manual_matching_gps_flow}
\end{figure}

This pseudocode is divided in 4 steps. Firstly, we assume that the absolute pose comes from a GPS source, however this is not the only option. Usually, the camera sensor records at 24 or 30 FPS, however the GPS can record anywhere between 5 to 100 updates per second. We need to synchronize the data by extrapolating or interpolating them missing points according to the RGB frequency. Then, we'll compute the Optical Flow of the video, using any off the shelf algorithm, such as \cite{sun2018pwc}. Plotting the mean magnitude of the optical flow per frame over all frames, alongside with the speed of the vehicle from GPS data results in a figure similar to Figure \ref{fig:manual_matching_gps_flow}.

What the Figure shows is that there is a strong correlation between optical flow and the speed of the vehicle. Most of the times when the magnitude of the flow is constant, the speed is also constant. When abrupt changes happen, such as acceleration or direction change, both signals will have spikes, but not necessarily in the same direction of the magnitude. We have manually annotated "hot" paths, where the synchroization should be done. We can see the strong correlation between the red and the green lines the signals were denoised using a median filter or similar signal processing techniques. Having an algorithm that matches these correlated matches will give us an offset. Finally, we will keep only the maximum intersection between the two newly synchronized data sources.

The next algorithm focuses on computing dense Depth maps from video sources and scaling them to match the absolute pose coordinate system.

\begin{algorithm}
\caption{Computing and scaling Depth from RGB + Pose}
\label{alg:alg-depth-scale}
\begin{algorithmic}[1]
\State $ Disparity, RP \gets \Call{DepthFromRGB}{RGB} $
\State $ ScaledRP \gets \Call{RelativeFromAbsolute}{Pose} $
\State $ Scale \gets \Call{FindScale}{RGB, ScaledRP, RP} $
\State $ ScaledDepth \gets \Call{ApplyScaling}{Scale, Disparity} $
\State $ \Return ScaledDepth $
\end{algorithmic}
\end{algorithm}

We can pick any off the shelf unsupervised depth estimation network, such as \cite{zhou2017unsupervised}. This method is able to take as input, once trained, any frame of a video and return an unscaled disparity map. It can also take any pair of two frames and return an unscaled relative pose between the two of them. From the set of synchronized absolute poses, having followed Algorithm \ref{alg:sync-rgb-abspose}, we can compute another set of scaled relative poses between each frame. Having access to these two sets of relative poses, we can compute the per frame scaling factor that can be applied to each disparity map. One way to do this is to apply the view synthesis operator with both sets of relative poses such that we match and minimize the difference between them. It should be noted that we don't really need to apply the last step to train the Pose2RGBD network, since we already have our 3 ingredients: RGB, Absolute Pose and (Unscaled) Depth. This is indeed the case for our real world dataset, where we synchronized the GPS with the video for all the flights, however, we only computed the unscaled depth maps.

Finally, this process can be done in a completly unsupervised fashion by composing multiple unscaled relative poses together in order to create a pseudo absolute pose, relative to a key frame. This is a well studied problem and the key challenges is that drifts will appear caused by accumulated errors, however these errors will be consistent with the predicted unscaled depth maps, therefore the dataset is synchronized. Bundle adjustment techniques can be applied to fix this as well as using more robust methods, like Structure from Motion and explicit 3D models.

\subsection{Proposed Datasets}

For the purpose of this paper, we have created two datasets, one based on the CARLA simulator \cite{Dosovitskiy17} and one taken by flying an UAV over the campus of an university and then applying the steps previously described in Algorithms \ref{alg:sync-rgb-abspose} and \ref{alg:alg-depth-scale}.

\subsubsection{Synthetic Dataset}

\begin{figure}[h]
\begin{center}
   \includegraphics[width=1\linewidth]{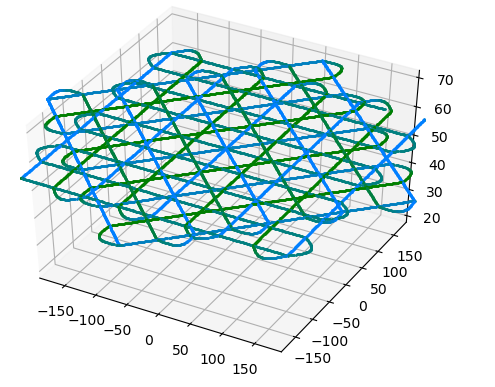}
\end{center}
   \caption{Synthetic Dataset flight path}
\label{fig:synthetic_dataset_flight_path}
\end{figure}

The synthetic dataset contains 11,085 RGBD images and Absolute Poses over a surface of about 300x300 meters. The images are taken at a resolution of 854x854. The altitude is kept at a constant of about 55m, while all the 3 rotation angles of the camera are varied with a random factor, however it always looks towards the city, similarly to how an UAV would fly. The purpose of this dataset is to show that, given perfect information, the network is able to learn sufficient information in order to compute an internal 3D representation over a relatively large surface. In Figure \ref{fig:synthetic_dataset_flight_path} we can observe the flight pattern used to export the dataset.

\subsubsection{Real World Dataset}

\begin{figure}[h]
\begin{center}
   \includegraphics[width=1\linewidth]{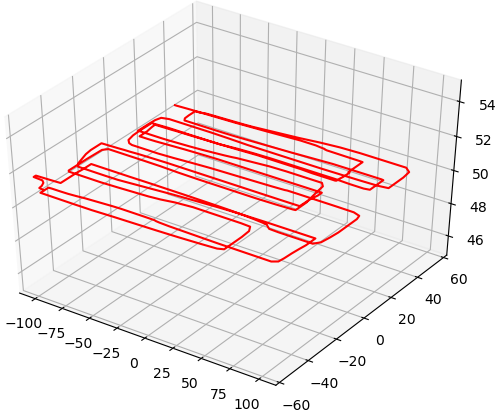}
\end{center}
   \caption{Real Dataset flight path}
\label{fig:real_dataset_flight_path}
\end{figure}

The real world dataset is constructed by doing an UAV flight over the campus of an university over a range of 260x150m. The recording is 12 minutes long and is recorded at 4K@24FPS. The dataset is sampled such that the altitude is constant at about 50m above the ground, as this is the desired setup, thus removing starting/ending frames where the UAV takes off and lands. In total, we get 15,605 usable RGB frames. The flight is done such that the gimbal is kept at almost the same angle, pointing towards the ground while the drone surveys in a U-shaped path, as can be seen in Figure \ref{fig:real_dataset_flight_path}. The RGB frames are then synchronized with the raw GPS log, as described in Algorithm \ref{alg:sync-rgb-abspose}. Then, we train an off-the-shelf unsupervised depth estimation network on these frames and export the depths map for each RGB frame, as described in Algorithm \ref{alg:alg-depth-scale}, however we only keep the unscaled version of the maps, so we have no correspondence between meters and the resulted values.

While the pattern is not as nice as the synthetic case, we can see that the flights try to maintain a preset route. In the following section, where we present the results of the Pose2RGBD network over these datasets, the first dataset will be called Synthetic, while the real dataset will be called Real.

\begin{figure}[h]
\begin{center}
   \includegraphics[width=1\linewidth]{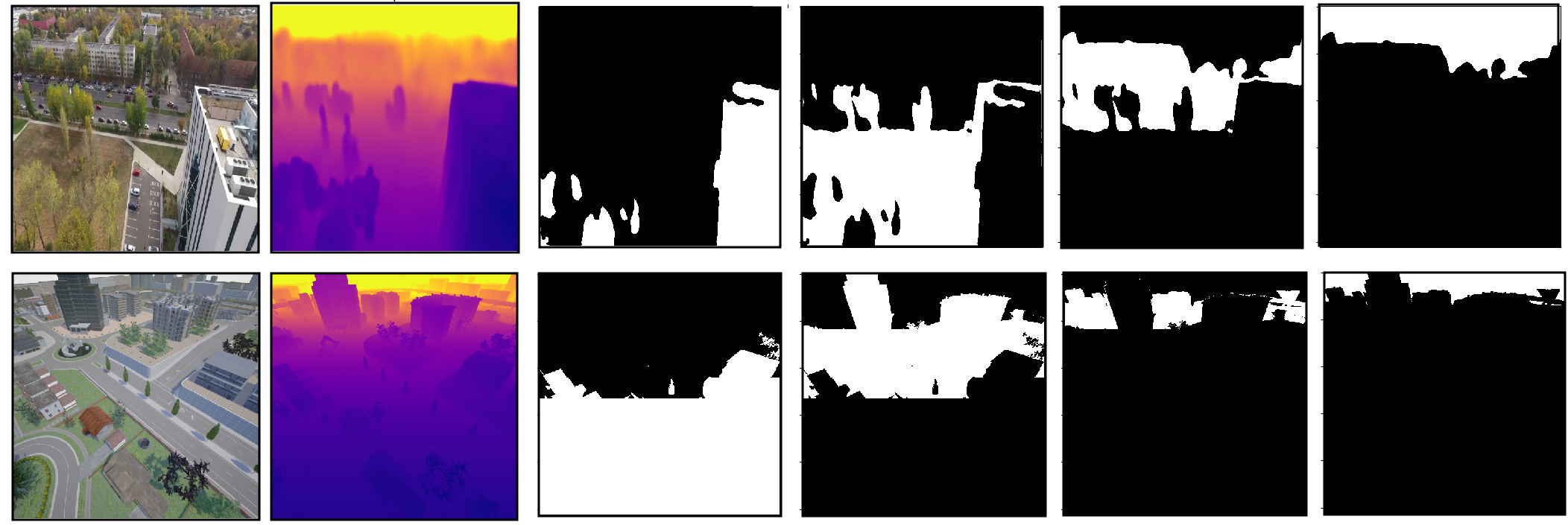}
\end{center}
   \caption{Samples from the two datasets alongside with a slicing of 4. Top row: Real dataset. Bottom row: Synthetic dataset.}
\label{fig:samples_real_synthetic_depthslices}
\end{figure}

In Figure \ref{fig:samples_real_synthetic_depthslices} we present two samples from the real and synthetic datasets using a depth slicing of 4. We observe how the slicing divides the scene in multiple regions, obtaining a soft segmentation over the scene directly from the depth maps.

\section{Experiments and results}

The two networks, which we'll call Base and Slice, as described in Figure \ref{fig:model-architecture}, were trained on the two datasets, Real and Synthetic. Most of the experiments were run on the Synthetic dataset and only a small subset of them were also redone on the Real dataset. Two input types were used, Translation + Euler angles, which will be called 6DoF and Translation + Quaternions, which will be called 6DoF-Quat. Depth errors are computed as mean pixel in meters for the Synthetic dataset. However, for the Real dataset, since the depth is not normalized, it has no real world significance. For the RGB predictions, we use the mean absolute pixel error in the range $[0 : 255]$. Since these two are strongly correlated, as we'll see, we can use this metric as a proxy for the quality of the depth map as well. All the inputs and outputs were normalized in the range of $[-1 : 1]$ and all the outputs had hyperbolic tangent as the final activation function. We also tried changing the normalization to $[0 : 1]$ or removing the activation function for the final layer, however the results were very similar (~98\% relative errors), so, for consistency, we kept the same setup everywhere. The train/validation splits are done by randomizing the frames' order and doing a 0.8/0.2 split. All outputs have a resolution of 512x512, regardless of the Dataset, in order to use the same models. This means that, even if the Real dataset has a 9:16 aspect ratio, we resize it to 1:1, which loses some realism, but in this way we can compare the results to the Synthetic dataset directly. All the models were trained using the PyTorch 1.3 \cite{paszke2017automatic} framework using the NVIDIA Tesla P100, up to 100 epochs and optimized using the AdamW optimizer \cite{loshchilov2017fixing}, a fixed learning rate of 0.01 and no hyperparameter tuning.

\begin{table}[h]
\begin{center}
\begin{tabular}{|l|l|l|l|l|}
\hline
Input & Output & RGB (px) & Depth (m) \\
\hline\hline
6DoF & Depth & n/a & 8.2 \\
6DoF & RGB & 20.96 & n/a \\
6DoF & RGBD & 21.45 & 8.85 \\
6DoF-Quat & Depth & n/a & \textbf{8.02} \\
6DoF-Quat & RGB & \textbf{20.78} & n/a \\
6DoF-Quat & RGBD & 21.13 & 8.41 \\
\hline
\end{tabular}
\end{center}
\caption{6DoF vs 6DoF-Quat results for the Base Model on the Synthetic Dataset.}
\label{table:6dof_vs_6dof-quat}
\end{table}

What we observe in Table \ref{table:6dof_vs_6dof-quat} is that the Quaternion representation outperforms the regular Euler angles representation in all the experiments, which is why they are used for all subsequent experiments. We also observe that, while there is a small drop in performance to predict both RGB and Depth, the relative performance loss is just of 1.65\%, respectively 4.64\%.

We then move to predicting both RGBD as well as Depth Slices, as described in Section \ref{sec:proposed-method}.

\begin{table}[h]
\begin{center}
\begin{tabular}{|l|l|l|l|l|}
\hline
Model & Output & RGB (px) & Depth (m) \\
\hline\hline
Base & Depth & n/a & 8.02 \\
Base & RGBD & 21.13 & 8.41 \\
Slice-10 & Depth & n/a & 7.23 \\
Slice-10 & RGBD & 20.71 & 7.01 \\
Slice-32 & Depth & n/a & \textbf{6.31} \\
Slice-32 & RGBD & 20.39 & 6.97 \\
Slice-64 & Depth & n/a & 6.64 \\
Slice-64 & RGBD & \textbf{20.37} & 7.1 \\
\hline
\end{tabular}
\end{center}
\caption{Base vs Slice models on the Synthetic dataset. The input is 6Dof-Quat, so it is omitted.}
\label{table:base_vs_slice_models_synthetic}
\end{table}

In Table \ref{table:base_vs_slice_models_synthetic} we observe how using the updated model with depth slices improves the quality of the results significantly. We observe, however, that blindly increasing the number of slices can actually decrease the performance, thus this is a hyperparameter which must be tuned accordingly. Some sceneries might perform better with a small number (e.g. indoor), while others might benefit with a larger number (e.g. aerial images).

\begin{table}[h]
\begin{center}
\begin{tabular}{|l|l|l|l|l|}
\hline
Model & Output & RGB (px) & Depth* \\
\hline\hline
Base & Depth & n/a & 0.01874 \\
Base & RGBD & 24.91 & 0.01915 \\
Slice-10 & Depth & n/a & 0.0147 \\
Slice-10 & RGBD & \textbf{24.82} & \textbf{0.01449} \\
\hline
\end{tabular}
\end{center}
\caption{Base vs Slice models on the Real dataset. The input is 6Dof-Quat, so it is omitted. *Note: Depth is unscaled.}
\label{table:base_vs_slice_models_real}
\end{table}

Table \ref{table:base_vs_slice_models_real} presents the results for the Real dataset. The Depth error is based on the internal scale of the method that was used to generate the labels, so it has no real world interpretation. However, we can correlate it to the pixel error of the RGB results, which can act as a proxy for fidelity. Using the ratio between RGB and Depth error in Table \ref{table:base_vs_slice_models_synthetic} can also give us a rough estimation of the actual depth error.

\begin{table*}[h]
\begin{center}
\begin{tabular}{|l|l|l|l|l|l|l|l|l|l|}
\hline
\multirow{2}{*}{Model} & \multirow{2}{*}{parameters} & \multirow{2}{*}{Batch Size} & \multirow{2}{*}{RAM (MB)} & \multicolumn{3}{|c|}{FPS} \\
& & & & CPU & GPU1 & GPU2 \\
\hline\hline
\multirow{3}{*}{Base} & \multirow{3}{*}{2,928,673} & 1 & 31.34 & 32.85 & 245.75 & 357.74 \\
& & 5 & 86.13 & 41.04 & 322.84 & 1080.73 \\
& & 10 & 124.14 & 37.16 & 336.15 & 1314.48 \\
\hline
\multirow{3}{*}{Slice-10} & \multirow{3}{*}{2,975,633} & 1 & 128.78 & 7.85 & 63.67 & 122.38 \\
& & 5 & 320.39 & 7.58 & 78.64 & 330.89 \\
& & 10 & 924.69 & 6.92 & 80.41 & 410.23 \\
\hline
\multirow{3}{*}{Slice-32} & \multirow{3}{*}{2,987,601} & 1 & 151.63 & 7.63 & 60.80 & 120.67 \\
& & 5 & 579.22 & 6.74 & 74.43 & 302.26 \\
& & 10 & 1258.21 & 6.67 & 76.66 & 365.44 \\
\hline
\multirow{3}{*}{Slice-64} & \multirow{3}{*}{3,005,009} & 1 & 164.69 & 7.04 & 56.86 & 117.71 \\
& & 5 & 655.28 & 6.49 & 68.87 & 264.56 \\
& & 10 & 1446.07 & 6.50 & 65.24 & 167.27 \\
\hline
\end{tabular}
\end{center}
\caption{Inference performance between multiple models. Experiments ran on Intel i7-6700HQ, NVIDIA GTX 960m and NVIDIA V100.}
\label{table:ablation_study_performance}
\end{table*}

Table \ref{table:ablation_study_performance} presents a study about performance in space and time on 3 different processing units. Inputs to the networks are represented as 6Dof-Quat and outputs are of shape 512x512x1 (Depth). The FPS is computed by multiplying the inference time to the batch size. In theory, we could compute, using a batch of 10, the current frame and 9 frames ahead, given a known trajectory, so the term FPS makes sense here. The experiments for each configuration were run 100 times and the results were averaged. The observed RAM (or VRAM for GPUs) was taken by computing the difference of memory usage before and after a forward pass. Since the models are so shallow, we observe very fast inference times, while still using little memory. Pipelining multiple consecutive frames also brings a benefit, since the model can make future predictions ahead of time and then just feed them with no computation cost. The number of slices should be treated as a hyperparameter, that is dependant on the type of data since it imposes a penalty hit if increased too much.

\begin{figure}[H]
\begin{center}
  \includegraphics[width=1\linewidth]{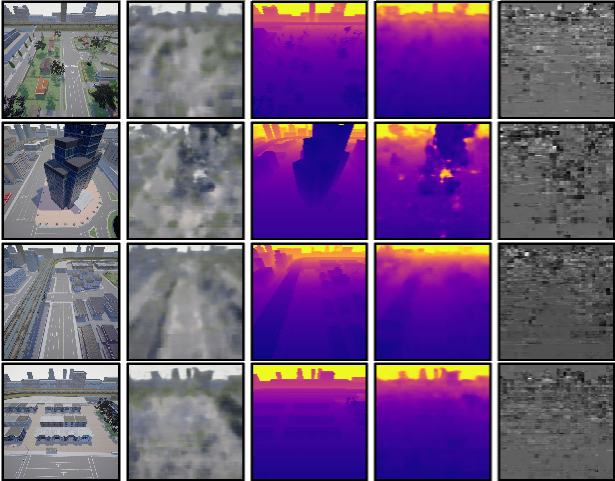} \\
  \vspace{0.2cm}
  \includegraphics[width=1\linewidth]{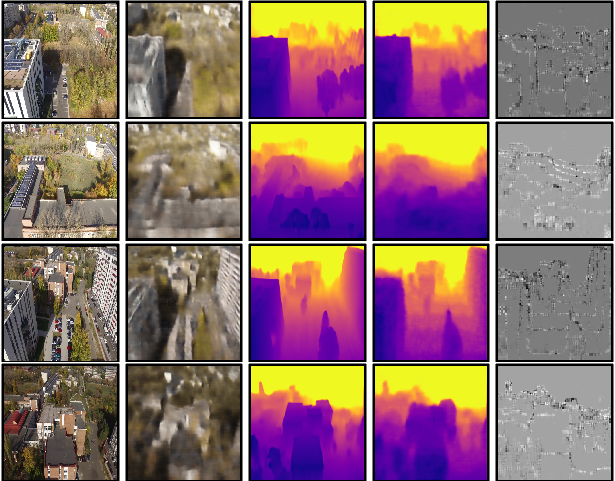}
\end{center}
  \caption{Qualitative results from the Synthetic (top 4 rows) and Real (bottom 4 rows) datasets.}
\label{fig:qualitative_results_synthetic_and_real}
\end{figure}

We end this section by providing a few qualitative results from the Synthetic and Real datasets, from the best performing models in Figure \ref{fig:qualitative_results_synthetic_and_real}. The first and third columns are ground truth RGB and Depths, while the second and forth columns are predicted maps. The fifth column is the confidence map, computed from summing each channel of the predicted depth slices, as described in Section \ref{sec:proposed-method}. Black pixels represent low confidence, white pixels are pixels predicted by multiple maps and gray pixels are confident pixels, predicted by only one map. Ideally, the whole picture should look gray.

\section{Conclusions and further work}
% \vspace{-10mm}
We have proposed a technique that implicitly learns a high level representation of a scene, using the correlation between RGB, Depth and Absolute Pose signals. Upon training, these models can be used as a neural renderer to produce novel RGBD images for a given pose input, even in unseen scenes. While the results are not of a very high quality, it should be noted that the input itself has no redundancy and is very low dimensional, so the network has to learn a direct mapping between position and the high dimensional visual space.

These networks resemble very much generative networks, such as GANs or VAEs, so adding an adversarial training or trying to minimize the divergence between the input distribution and the output reconstruction are very obvious next steps that should improve the results.

We could also add other helpful cost functions, such as consistency losses between nearby poses, as done by \cite{zhang2019unsupervised} should add a significant benefit to the results.

{\small
\bibliographystyle{ieee_fullname}
\bibliography{egbib}
}

\end{document}